\pdfoutput=1

\documentclass[11pt]{article}

\usepackage[]{acl}
\usepackage{booktabs}
\usepackage{times}
\usepackage{latexsym}
\usepackage[utf8]{inputenc}
\usepackage{CJKutf8}
\usepackage{times}
\usepackage{latexsym}
\usepackage{amsmath}
\usepackage{url}
\usepackage{booktabs}
\usepackage{adjustbox}
\usepackage[utf8]{inputenc}
\usepackage{CJK}
\usepackage{tabularx}
\usepackage{multirow}
\usepackage{graphicx} 
\usepackage{amsfonts} 
\newcommand{\kaiti}[1]{\begin{CJK*}{UTF8}{gkai} #1 \end{CJK*}}

\usepackage[T1]{fontenc}

\usepackage[utf8]{inputenc}

\usepackage{microtype}

\usepackage{inconsolata}

\usepackage{graphicx}

\title{Contrastive Token Learning with Similarity Decay for Repetition Suppression in Machine Translation}

\author{
    \textbf{Huangyu Dai\textsuperscript{1,\thanks{Equal Contribution.}}}  
    \textbf{Ben Chen\textsuperscript{1,\footnotemark[1],  \thanks{Corresponding Author.}}}  
    \textbf{Kaidi Chen\textsuperscript{1,2}}  
    \\
    \textbf{Ying Han\textsuperscript{3}}  
    \textbf{Zihan Liang\textsuperscript{1}} 
    \textbf{Wen Jiang\textsuperscript{1}}   
    \\
    \textsuperscript{1}{Alibaba Group, Hangzhou, China}\\
    \textsuperscript{2}{Northwestern Polytechnical University, School of Automation, Xi'an, China} \\
    \textsuperscript{3}{Zhejiang Gongshang University, School of Foreign Languages, Zhejiang, China} \\
}

\begin{document}
\maketitle
\begin{abstract}

For crosslingual conversation and trade, Neural Machine Translation (NMT) is pivotal yet faces persistent challenges with monotony and repetition in generated content. Traditional solutions that rely on penalizing text redundancy or token reoccurrence have shown limited efficacy, particularly for lengthy article and e-commerce descriptions with inherent redundancy, even with the advent of Large Language Models (LLMs). This paper investigates the underlying causes of textual repetition through the lens of information entropy, attributing the phenomenon to the elevated uncertainty within the input text. To address this, a novel algorithm named Contrastive Token Learning with Similarity Decay (CTSD) is introduced, which modulates the suppression of tokens dynamically, informed by varying attention weights and inter-token distances. Furthermore, an e-commerce dataset comprised of title texts of online real items is compiled and released susceptible to hallucination translations to benchmark the algorithm. Extensive evaluations demonstrate that CTSD significantly outperforms existing approaches in precision and generalizability. Additional online A/B testing underscores its practical value, showing marked improvements in user engagement and conversion. Notably, this method has been implemented with full traffic on eight multilingual sites of alibaba.com, the largest B2B e-commerce platform in the world.
\end{abstract}

\section{Introduction}

In recent years, the synergy of neural networks coupled with the increasing scale of parallel corpora has significantly propelled Neural Machine Translation (NMT) forward \citep{10.1162/tacl_a_00343, costa2022no}. Notably, the sophisticated reasoning abilities and specialized knowledge acquired by Large Language Models (LLMs) \citep{touvron2023llama, qwen} further contribute modern NMT systems towards achieving near-human-level performance \citep{lin2022automatic, zhu2023multilingual}. However, the reliability of NMT in delivering accurate and coherent translations remains unstable, often encountering unexpected errors such as omissions or nonsensical outputs. This challenge persists across the spectrum, especially for complex textual materials like repetition-prone articles and e-commerce descriptions.

\begin{table*}[htbp]
\centering
\begin{adjustbox}{max width=\textwidth}
\renewcommand{\arraystretch}{0.95}
\begin{tabular}{lcl}
\toprule
\textbf{Model} & \textbf{Type} & \textbf{Sentence} \\

\midrule
\multirow{3}{*}{NLLB-1.3B} &  src\_t &     Baseball cap Manufacturer Custom plain Baseball hat Embroidered baseball cap for men  \\
 &  tran\_t  &    Baseballkappe Hersteller, Baseballhut, Stück für Stück, Baseballhut, Stück für Stück, Stück für Stück, ...   \\
 &  opt\_t  &  Baseballmütze Hersteller individuelle schlichte Baseballmütze bestickte Baseballmütze für Herren \\  \midrule
 \multirow{3}{*}{mBART-large} &  src\_t &  1.8 Ton Mini Excavator Crawler Excavator Mini Bagger Cheap Price With Ce For Sale Epa Ce Mini Excavator\\
 &  tran\_t  &  1,8 Tonnen Mini Bagger Bagger Bagger Bagger Bagger Bagger Bagger ... \\
 &  opt\_t  &   1,8 Tonnen Mini Bagger Mini Bagger Preis mit Ce Zum Verkauf Epa Ce Mini Bagger \\  \midrule
 \multirow{3}{*}{LLaMA2-7B} &  src\_t &   4 in 1 modern rotating multi game billiard pool table 7ft with air hockey 4 in 1 pool table 4 in 1 table game  \\
 &  tran\_t  &   4-in-1 moderne rotateürende multi-Spiel-Billard-Pool-Tisch 4-in-1-Tisch-Spiel, 4-in-1-Tisch-Spiel... \\
 &  opt\_t  &    4-in-1 moderner drehbarer multi-spiel-billardtisch 7-fuss mit air-hockey 4-in-1-pooltisch 4-in-1-tischspiel  \\  \midrule
 \multirow{3}{*}{Qwen-7B} &  src\_t &  Excavator Machine electric Hydraulic Mini Small Micro Crawler Bagger Digger Mini Excavators \\
 &  tran\_t  &   Abbaumaschine Elektrohydraulische Kleinmodell Mikro-Krabbenwerfer Mini-Bagger\kaiti{迷你}\kaiti{挖掘机挖掘机} ... \\ 
 &  opt\_t  &   Bagger Abbauger Elektro-hydraulik klein Mikro-Crawler-abbaugern minibaggerminibagger \\
 
\bottomrule
\end{tabular}
\end{adjustbox}
\caption{Examples of repetition generation in NMT. \text{Src\_t}, \text{tran\_t}, and \text{opt\_t} are the abbreviations of source texts, translated texts with the original model, and optimized texts with the additional CTSD method.}
\label{table:your_label}
\vspace{-1em}
\end{table*}


Typical NMT problems, commonly referred to as "hallucinations", can be categorized into two main types \citep{dale2022detecting, guerreiro2023looking}. The first involves the repetition of words or sentences, known as "oscillations", while the second pertains to the generation of content not supported by the source, termed "largely fluent". Of these two types, "oscillations" are particularly intolerable for leading to repetition with low coherence and accuracy, making NMT limited for multiple applications \citep{ji2023survey, guerreiro2023hallucinations}. Consequently, addressing oscillation (repetition generation) has emerged as a primary focus in current research, which is vital for improving reliability and usability in complex scenarios.

Previous methods mainly employed two strategies to suppress repetition generation. The first is the direct strategy interventions during the inference stage, such as n-gram not repeat, Contrastive Search (CS) \citep{su2022contrastive}, and Penalized Sampling (PS) \citep{keskar2019ctrl}. These techniques focus on preventing repeated tokens to eliminate oscillations. However, they would disrupt the token distribution of output, leading to other errors. Consequently, recent methods focus on designing training objectives during the model training stage to better address hallucination problems \citep{welleck2020neural, su2022contrastive, jiang2022simple}. Yet, these training objectives do not adequately explore the root of oscillations in transformer-based models, often using a direct intervention way to suppress tokens that have appeared. Although they can effectively suppress word or sentence repetition, they also lead to a lower coherence and accuracy in translation \citep{post2018call, wan-etal-2022-unite}. 

With the emergency of LLMs, reinforcement learning (RL) methods such as Proximal Policy Optimization (PPO) and Direct Preference Optimization (DPO) have surfaced as a new strategy for reducing hallucinations \citep{schulman2017proximal, rafailov2024direct}. Through training LLMs with preference data, these methods generate outputs close to human expectations, effectively lowering the chance of hallucinations.

In this paper, we conduct an in-depth exploration of the fundamental reasons underlying textual repetition in machine translation, primarily utilizing the concept of information entropy. Our research reveals that this phenomenon largely stems from increased levels of uncertainty present within the input text. Repetitive token generation occurs because the information from previously generated tokens does not provide additional value (information entropy). To effectively deal with this issue, we propose an innovative algorithm, which we term "Contrastive Token Learning with Similarity Decay" (CTSD). This innovative approach aims to dynamically adjust the suppression of tokens by analyzing the attention differences between different output tokens’ embeddings and the distance between the inner tokens, thereby enhancing the accuracy and stability of the output. Meanwhile, our method can be applied to both specialized translation models and LLM without additional data preparation. The results show that our method can effectively improve the performance of models in translation tasks and prevent oscillation hallucinations. Compared with Contrastive Token Learning (CT), CTSD achieves improvements by 1\% to 10\% in translation quality on both the FLORES-200 and our proprietary e-commerce datasets.

In addition, a comprehensive evaluation by experts reveals that CTSD exceeds current methods in both precision and generalizability. Online A/B tests further highlight its practicality, as evidenced by substantial gains in user engagement with higher click-through and conversion rates and final gross merchandise volume. Importantly, this method has been successfully deployed across eight multilingual websites with full traffic of alibaba.com, the world's largest B2B e-commerce platform.

\begin{figure*}[htbp]
    \centering
    \includegraphics[width=1\linewidth]{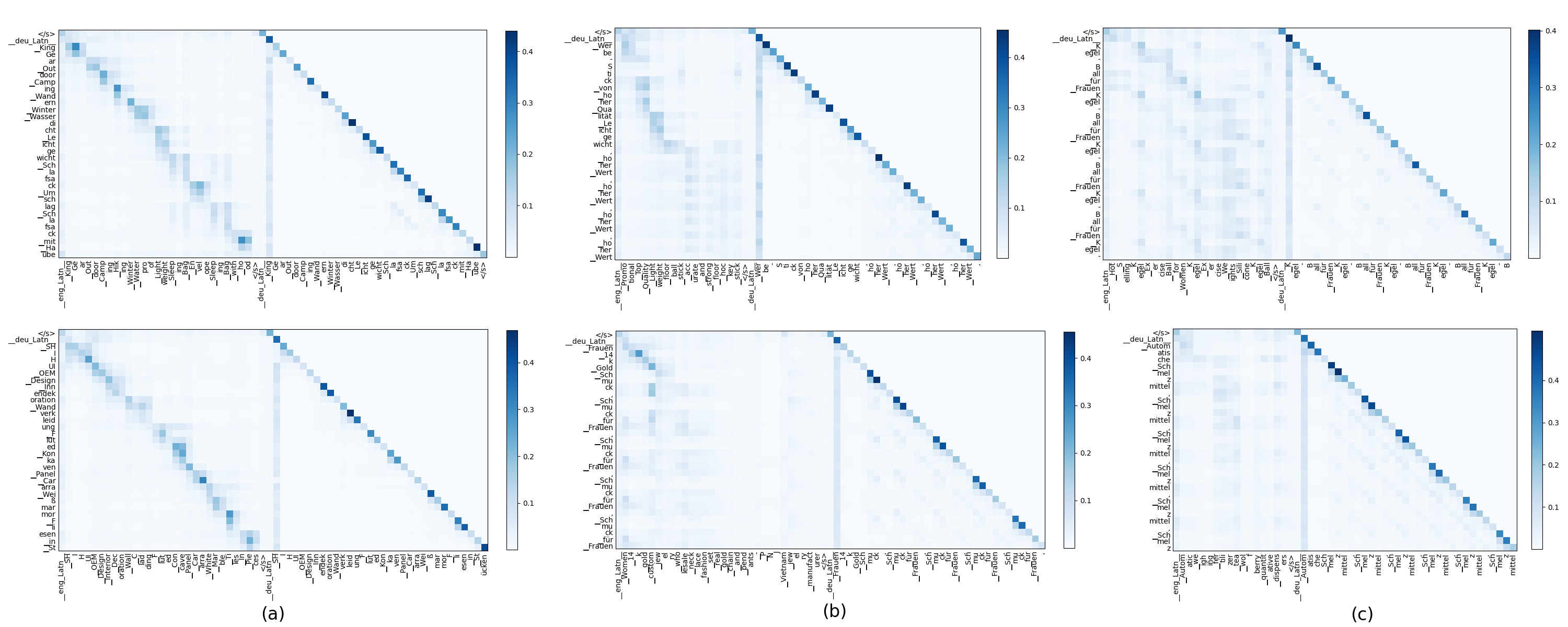}
    \caption{ALTI+ results for En-De translation examples. (a) normal result. (b) middle appearing repetition result, and (c) total repetition result. The contribution values of all tokens in each row have been normalized.}
    \label{fig:alti}
\vspace{-0.5em}
\end{figure*}

\section{Related Work}

\subsection{Multilingual Neural Machine Translation}

Multilingual NMT has advanced significantly from its early focus on two-language systems. The pioneering work of Dong et al. expanded NMT into a one-to-many framework by sharing encoders across four language pairs \citep{dong2015multi}. This development sparked a surge in research on NMT systems capable of handling multiple languages \citep{johnson2017google,  chu2019multilingual, yang-etal-2021-multilingual}. At first, the research was mainly focused on improving multilingual NMT's capabilities on rich-resource languages through designed specific components and more diverse training data \citep{escolano2020training, fan2021beyond, 10.1162/tacl_a_00343}. Now, more research has turned to low-resource languages. Tars et al. improved the capabilities of low-resource languages by simultaneously training different languages of the same language family \citep{tars2021extremely}. Pan enhanced the translation quality of non-English language directions through data augmentation and contrastive learning \citep{pan2021contrastive}. NLLB Team developed a conditional compute model based on a Sparsely Gated Mixture of Experts to improve low-resource language translation quality \citep{costa2022no}.

With the increasing scale of parameters and training corpus, LLMs (GPT-3, BLOOM, and LLaMA included in open source models \citep{brown2020language, workshop2022bloom, touvron2023llama}, ChatGPT, GPT-4 and Claude included in closed source models \citep{ openai2022chatgpt, achiam2023gpt, Claude}) have gained unexpected complex reasoning and emergent abilities in the face of unseen tasks, enabling it to handle various tasks, like text summarization, QA system, and free dialogue \citep{wei2022emergent}. However, the professional evaluation found that many large models still cannot surpass state-of-the-art translation engines like NLLB and Google Translate in professional translation \citep{zhu2023multilingual}. The recently emerged translation-specialized LLMs attempt to conduct more specialized data training to reduce this gap \citep{xu2023paradigm, chen2024general2specialized}. However, they still inevitably generate repetition when faced with a complex text translation.

\subsection{Repetition Suppression}

With the advent of LLMs, repetition suppression methods have received significant attention. Currently, there are two mainstream types of methods: decoding methods and training methods. Decoding methods initially gained popularity because no further tuning is needed. Commonly used methods include PS and CS. Keskar et al. implemented PS, using a temperature coefficient to reduce the likelihood of historical tokens, reducing the probability of producing oscillatory hallucinations \citep{keskar2019ctrl}. Su et al. proposed CS to suppress historically generated tokens by computing the cosine similarity between the embedding of historical tokens and the current token \citep{su2022contrastive}.

While decoding methods successfully suppress oscillation hallucination, they face challenges of reduced generation quality and increased inference cost. Therefore, research is shifting towards designing training objectives for more accurate and stable translations. Welleck et al. proposed unlikelihood training (UL), suppressing repetition through unlikelihood loss \citep{welleck2020neural}. Su et al. adopted contrastive training, emphasizing distinctions between different tokens to prevent monotonous repetition \citep{su2022contrastive}. Jiang et al. introduced the CT loss, which selectively suppresses tokens on a negative token list without impacting irrelevant tokens \citep{jiang2022simple}. CT has been theoretically proven to be advantageous over traditional cross-entropy (CE) and UL loss and emerged as the most effective algorithm of oscillations suppression to date \citep{sun2023contrastive, guan2023generating}.

\section{Methodology}

\subsection{Hallucination Analysis}

A well-known theoretical analysis of the repetition problem in text generation simplifies predicting the next word into a first-order Markov chain \citep{fu2021theoretical}. It assumes that the currently generated token is only affected by the same token generated at the previous moment. Under this assumption, the entire generation sequence forms a directed cycle when the model generates a word that has already been generated. For the sentence "I like it and guess he knows I like it because ...", this theory insists that the second generation of "I like it" is mainly affected by the former so that the next predicted token is most likely "and". Methods like CT and PS employ this idea to prevent directed cycles and reduce repetitive text generation.

However, this theory ignores the input text and previously translated text, contrary to the model based on the cross-attention mechanism. Consider the title of an item: "Best Selling New Arrival Outdoor Shapewear Dress Women's Dresses Built-in Shapewear Maxi Dress". Global-level suppression of repetitive generation can lead to the replacement of "shapewear" and "dress" in the latter part of the title translation with other words, thus deviating from the original meaning. To more accurately analyze the impact of each former token on the next predicted word, we perform visual analysis through the ALTI+ method \citep{ferrando2022towards}. ALTI+ calculates the contribution of each previous token to the generation of the current token by comparing the Manhattan distance between the previous token and the newly generated token representation. For a comprehensive comparison, we show En-De translation with three columns: (a) normal result, (b) middle appearing repetition result, and (c) total repetition result, respectively, in Figure~\ref{fig:alti}.

We can distinctly observe that: 1) The generation of each token is primarily influenced by the input text and the nearest neighboring tokens, with tokens at relatively farther positions exerting minimal impact; 2) Identical repeated words are affected by tokens in the same position, and the longer the generated text, the weaker the influence of the corresponding tokens in the input text. This explains why global suppression of repeated words, although effective in suppressing repetition, leads to poorer translation outcomes, particularly in decoder-only LLMs, where a forcibly replaced repetitive token results in subsequent tokens deviating increasingly from the original meaning.

Here, we attempt to elucidate the underlying causes of text repetition generation from the perspective of information entropy. In the transformer mechanism, the generation of each token is influenced by all preceding tokens. Repetitive token generation occurs because the information from previously generated tokens does not provide additional value. For instance, when predicting the (\textit{n}+1)th token, the information from tokens 0 to (\textit{n}-1) is identical to that from tokens 0 to \textit{n}, causing the model to generate the \textit{n}th word repetitively.

\begin{figure}[htb]
    \centering
    \includegraphics[width=1\linewidth]{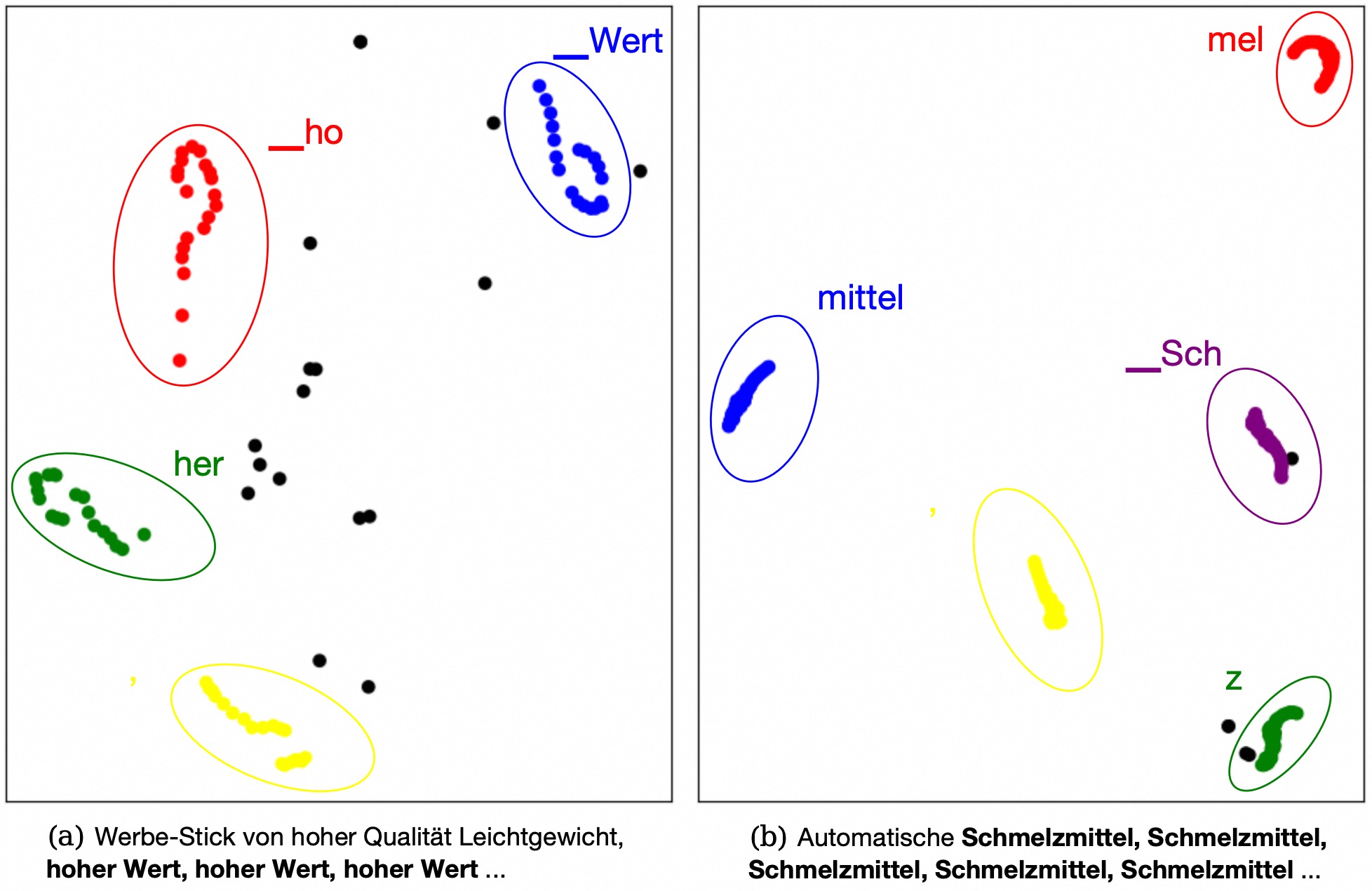}
    \caption{T-SNE results of different generated tokens. (a) middle appearing repetition result and (b) total repetition result.}
    \label{fig:tsne}
\vspace{-1em}
\end{figure}

To substantiate this, we have calculated the embeddings of each token and visualized it using T-SNE (shown in Figure~\ref{fig:tsne}). We can observe that the two-dimensional vectors of repeated tokens are clustered together, and two identical tokens generated continuously are closely together. The average cosine similarity of two adjacent "her" tokens in the first picture is 0.85, which is much higher than two adjacent different tokens. In the second picture, the cosine similarity of two adjacent "mittel" tokens is 0.94, but the similarity between two different tokens that appear before and after is only 0.33. This shows that when repeated tokens are generated, the generated text's information entropy remains unchanged. Therefore, to suppress repetitive generation while maintaining translation accuracy and stability, we should selectively focus on each former token and adaptively attend to the changes in information entropy with each token generation.

\begin{table*}[htb]
\centering
\begin{adjustbox}{max width=\textwidth}
\renewcommand{\arraystretch}{0.99}
\begin{tabular}{lcccccccccc}
\toprule
\textbf{Dataset} & \textbf{Model} & \textbf{Method} & \textbf{SacreBLEU↑} & \textbf{Rouge-L↑} & \textbf{COMET↑} & \textbf{rep-2↓} & \textbf{rep-3↓} & \textbf{rep-w↓} & \textbf{rep-r↓} & \textbf{div↑}  \\
\midrule

\multirow{15}{*}{FLORES-200} & Ground Truth        & -            & -               & -              & -           & 0.43            & 0.1           & 0.03           & 0.01            & 1.00      \\
\cmidrule(lr){2-11}
  & \multirow{3}{*}{NLLB-1.3B}   & CE         & 31.77             & 0.558               & 0.840              & 0.69           & 0.22            & 0.04            & 0.02            & 0.99       \\

&  & CT     & 30.51             & 0.547               & 0.841               & 0.33           & \textbf{0.10}            & 0.02           & 0.01            & 1.00            \\
&  & CTSD          & \textbf{32.14}             & \textbf{0.561}               & \textbf{0.843}               & \textbf{0.53}           & 0.13            & \textbf{0.03}           & \textbf{0.01}         & \textbf{1.00}        \\
\cmidrule(lr){2-11}
& \multirow{3}{*}{mBART-large} & CE          & 27.87             & 0.499               & 0.818               & 0.56           & 0.12            & 0.04           & 0.01            & 0.99                   \\
 & & CT          & 27.27             & 0.499               & 0.816               & 0.58           & 0.21            & 0.03            & 0.01           & 0.99                    \\
 & & CTSD          & \textbf{28.04}             & \textbf{0.508}               & \textbf{0.820}               & \textbf{0.56}           & \textbf{0.12}            & \textbf{0.03}            & \textbf{0.01}           & \textbf{0.99}                     \\
\cmidrule(lr){2-11}
& \multirow{3}{*}{LLaMA2-7B} & CE          & 19.65             & 0.439               & 0.826               & 4.29          & 3.84            & 0.05           & 0.02           & 0.88                   \\
 & & CT          &  19.42             &  0.439               & 0.827               & 0.90          & \textbf{0.24}            & 0.04            & 0.02           & 0.99                    \\
 & & CTSD          & \textbf{19.94}             & \textbf{0.443}              & \textbf{0.827}               & \textbf{0.86}           & 0.25            & \textbf{0.04}            & \textbf{0.02}           & \textbf{0.99}                     \\
\cmidrule(lr){2-11}
& \multirow{3}{*}{Qwen-7B}   & CE          & 18.99             & 0.426               & 0.776               & 0.26           & 0.12            & 0.02           & 0.01            & 1.00                   \\
 & & CT          & 19.53             & 0.435               & 0.780               & 0.28           & 0.07            & 0.02            & 0.01           & 1.00                    \\
 & & CTSD          & \textbf{19.72}             & \textbf{0.441}               & \textbf{0.784}               & \textbf{0.31}           & \textbf{0.09}           & \textbf{0.02}           & \textbf{0.01}           & \textbf{1.00}                     \\
\cmidrule(lr){2-11}
  &    GPT-3.5-Turbo         & -            & 3.86               & 0.065               & 0.735           & 0.46            & 0.12           & 0.03           & 0.01            & 0.99     \\
\cmidrule(lr){2-11}
 &    GPT-4-Turbo        & -            & 3.81               & 0.061               & 0.739           & 0.40            & 0.09           & 0.03           & 0.01            & 0.99      \\
\midrule
\multirow{12}{*}{E-Commerce}  & \multirow{3}{*}{NLLB-1.3B} 
& CE          & 6.71             & 0.178             & 0.575              & 36.17           & 37.21            & 0.13            & 0.10            & 0.24                   \\
& & CT         & 7.16             & 0.182               & 0.600               &0.82          & 0.21            & \textbf{0.05}           & 0.03            & 0.99                      \\
& & CTSD         & \textbf{7.59}             & \textbf{0.192}                & \textbf{0.602}               & \textbf{0.75}           & \textbf{0.19}            & \textbf{0.05}            & \textbf{0.02}           & \textbf{0.99}                  \\
\cmidrule(lr){2-11}
 & \multirow{3}{*}{mBART-large} & CE & 16.99             & 0.357             & 0.658              & 23.68           & 17.08            & 0.35            & 0.40            & 0.54                   \\
 & & CT          & 17.23             & 0.380               & 0.687               & 18.95          & 13.18            & 0.29           & 0.31            & 0.63                      \\
 & & CTSD          & \textbf{17.67}             & \textbf{0.391}               & \textbf{0.694}               & \textbf{12.66}           & \textbf{6.13}            & \textbf{0.29}            & \textbf{0.31}           & \textbf{0.79}                     \\
\cmidrule(lr){2-11}
 & \multirow{3}{*}{LLaMA2-7B} & CE & 19.06             & 0.436            & 0.747              & \textbf{0.59}           & \textbf{0.04}            & 0.06           & 0.02            & 0.99                   \\
 & & CT          & 20.72             & 0.455               & 0.753               & 0.82          & 0.14            & 0.06           & 0.02            & 0.99                      \\
 & & CTSD          & \textbf{21.11}             & \textbf{0.460}               & \textbf{0.757}               & 0.80           & 0.12            & \textbf{0.06}            & \textbf{0.02}           & \textbf{0.99}                     \\
\cmidrule(lr){2-11}
 & \multirow{3}{*}{Qwen-7B} & CE & 24.14             & 0.457             & 0.734              & 0.73           & 0.12            & 0.05            & 0.02            & 0.99                   \\
 & & CT          & 24.01             & 0.457               & 0.730               & 0.73          & 0.22            & 0.05           & 0.02           & 0.99                      \\
 & & CTSD          & \textbf{24.58}            & \textbf{0.462}               & \textbf{0.741}               & \textbf{0.73}           & \textbf{0.12}            & \textbf{0.05}            & \textbf{0.02}          & \textbf{0.99}                     \\
\cmidrule(lr){2-11}
  &    GPT-3.5-Turbo         & -            & 5.28               & 0.112               & 0.579           & 0.80            & 0.20           & 0.03           & 0.02            & 0.99     \\
\cmidrule(lr){2-11}
 &    GPT-4-Turbo        & -            & 3.84               & 0.081               & 0.582           & 0.57           & 0.08           & 0.02           & 0.01            & 0.99      \\
\bottomrule
\end{tabular}
\end{adjustbox}
\caption{Translation quality and repetition rate of NLLB-1.3B, mBART-large, LLaMA2-7B, and Qwen-7B models under different training methods and different datasets.}
\label{table:fulltable}
\vspace{-0.5em}
\end{table*}

\subsection{Learning-Based Solution}

Models trained with traditional CE loss are prone to hallucinations when facing out-of-domain data. To address the issue in translation models, several new training objectives (loss functions) have been designed to suppress negative tokens more effectively. Welleck et al. proposed unlikelihood training, specifically designed to penalize the likelihood of negative tokens \citep{welleck2020neural}. The token-level unlikelihood training objective (UL-T) at time step \(t\) is defined as:
\begin{equation}
\small
\begin{aligned}
\mathcal{L}_{UL}^t=-\sum_{y_t^{-} \in C^t} \log \left(1-p\left(y_t^{-} \mid \mathbf{y}_{<t}, \mathbf{x}\right)\right)
\end{aligned}
\end{equation}
\vspace{-1em}

where \(\mathbf{x}\) is the source text, \(\mathbf{y}_{<t}\) is the translation text generated before time step \(t\), \(C^t\) is the set of previous negative tokens at time step \(t\). This approach focuses on decreasing the probability of generating already-produced tokens, aiming to break the directed cycle observed in NMT models.

Additionally, contrastive learning loss \(\mathcal{L}_{CL}\) has been proposed as an effective training objective \citep{su2022contrastive}, which encourages models to learn isotropic token representations through a similarity penalty. The \(\mathcal{L}_{CL}\) is defined as:

\vspace{-1em}
\begin{equation}
\small
\begin{aligned}
\mathcal{L}_{CL}^{t} = \frac{1}{t-1} \sum_{i=1}^{t-1} \max\! \left\{0, \rho-\!s\left(h_{y_i}, h_{y_i}\right) + s\left(h_{y_i}, h_{y_{t-i}}\right)\right\}
\end{aligned}
\end{equation}
where \(\rho \in [-1, 1]\) is a pre-defined margin, \(h_{y_i}\) is the embedding of token \(y_i\), \(s\left(h_{y_i}, h_{y_{t-i}}\right)=({h_{y_i}^{\top} h_{y_j}})\setminus (\left\|h_{y_i}\right\| \cdot\left\|h_{y_{t-i}}\right\|)\) is cosine similarity. This loss function is designed to increase the distance between representations of distinct tokens, thereby creating a more discriminative and diverse model representation space.

Recently, CT loss is presented \citep{jiang2022simple} and the formulation for time step \(t\) is defined as:

\vspace{-1.5em}
\begin{equation}
\small
\begin{aligned}
\mathcal{L}_{CT}^t=\log \left(1+\sum_{y_t^{-} \in S_N^t} \exp \left(h_t^T W_{y_t^{-}} - h_t^T W_{y_t}\right)\right)
\end{aligned}
\end{equation}
where \(h^t\) is the hidden state, \(y_t\) means the positive token at time step \(t\). \(W_{y_t}\) denotes the embedding for token  \(y_t\), \(S_N^t\) is the set of the previous \(N\) tokens.

\begin{table}[htb]
\centering
\begin{adjustbox}{max width=\textwidth/2}
\begin{tabular}{lcccc}
\toprule
\textbf{Model} & \textbf{Dataset} & \textbf{Method} & \textbf{SacreBLEU↑}  &  \textbf{rep-2↓}  \\
\midrule
\multirow{4}{*}{Qwen-1.8B}   & \multirow{2}{*}{FLORES-200} 
& CE      & 5.79            & 1.88                 \\
& & CTSD      & \textbf{5.88}      & \textbf{0.19}   \\
\cmidrule(lr){2-5}
& \multirow{2}{*}{E-Commerce} 
 & CE     & 18.83          & 1.69   \\
 & & CTSD          & \textbf{19.00}    & \textbf{0.71}     \\
\midrule 
\multirow{4}{*}{Qwen-14B}   & \multirow{2}{*}{FLORES-200} 
& CE      & 21.47       & 0.32      \\
& & CTSD      & \textbf{21.80}     & \textbf{0.22}    \\
\cmidrule(lr){2-5}
& \multirow{2}{*}{E-Commerce} 
 & CE     & 26.42  & 1.42   \\
 & & CTSD          & \textbf{26.46}  & \textbf{0.72}    \\
\bottomrule
\end{tabular}
\end{adjustbox}
\caption{Translation quality and repetition rate of Qwen-1.8B and Qwen-14B models under different training methods and different datasets.}
\label{table:different scale model}
\end{table}

\begin{table*}[htb]
\centering
\begin{adjustbox}{max width=\textwidth}
\renewcommand{\arraystretch}{0.95}
\begin{tabular}{lcccccccccc}
\toprule
\textbf{Model} & \textbf{Method} & \textbf{SacreBLEU↑} & \textbf{Rouge-L↑} & \textbf{COMET↑} & \textbf{rep-2↓} & \textbf{rep-3↓} & \textbf{rep-w↓} & \textbf{rep-r↓} & \textbf{div↑} & \textbf{uniq-1↑} \\
\midrule
\multirow{8}{*}{NLLB-1.3B}   & -      & 0.71         & 0.098           & 0.280     & 93.29           & 92.65           & 0.91           & 0.95           & 0.00            & 7355            \\
\cmidrule(lr){2-11}
 & PS     & 6.47         & 0.173           & 0.571     & 3.69           & 3.54           & 0.03           & 0.02           & 0.89            & 13955            \\
 & CS          & 5.03         & 0.139           & 0.482     & \textbf{0.18}          & \textbf{0.04}            & \textbf{0.01}           & \textbf{0.00}            & \textbf{1.00}            &  \textbf{18378}            \\
\cmidrule(lr){2-11}
 & CE          & 6.71             & 0.178             & 0.575              & 36.17           & 37.21            & 0.13            & 0.10            & 0.24            & 13060            \\
 & UL-T          & 7.01             & 0.183               & 0.578              & 34.91           & 35.95            & 0.13            & 0.09            & 0.26           & 12547            \\
 & CL         & 6.84             & 0.181               & 0.578               & 26.51           & 27.23           & 0.11            & 0.08           & 0.38            & 12881           \\
 & CT          & 7.16             & 0.182               & 0.600               & 0.82          & 0.21            & 0.05           & 0.03            & 0.99            & 13670            \\
 & CTSD          & \textbf{7.59}             & \textbf{0.192}                & \textbf{0.602}               & 0.75           & 0.21            & 0.05               & 0.02            & 0.99         & 12658            \\
\midrule 
\multirow{8}{*}{Qwen-7B}   & -      & 4.14         & 0.133           & 0.599     & 5.66           & 5.22           & 0.04           & 0.03           & 0.85            & 14072            \\
\cmidrule(lr){2-11}
 & PS     & 3.29         & 0.113           & 0.593     & 0.97           & 0.39           & 0.03           & 0.02           & 0.98            & \textbf{14265}            \\
 & CS          & 4.09         & 0.135           & 0.599     & 2.84         & 2.19            & 0.04           & 0.03            & 0.93           &  13973            \\
\cmidrule(lr){2-11}
 & CE          & 24.14             & 0.457             & 0.734              & 0.73           & 0.12            & 0.05            & 0.02            & 0.99     & 11303            \\
 & UL-T          & 24.13             & 0.459               & 0.736              & 0.97           & 0.43            & 0.05            & 0.02            & 0.98           & 11252            \\
 & CL         & 24.55             & 0.460               & 0.740               & 0.74           & 0.12           & 0.05            & 0.02           & 0.99            & 11103           \\
 & CT          & 24.01             & 0.457               & 0.730               & 0.73          & 0.22            & 0.05           & 0.02           & 0.99          & 11227            \\
 & CTSD          & \textbf{24.58}             & \textbf{0.462}               & \textbf{0.741}               & \textbf{0.73}           & \textbf{0.12}            & \textbf{0.05}            & \textbf{0.02}          & \textbf{0.99}       & 11130            \\
\bottomrule
\end{tabular}
\end{adjustbox}
\caption{Translation quality and repetition rate of NLLB-1.3B and Qwen-7B under different repetition suppression methods during training or inference stage.}
\vspace{-0.5em}
\label{table:repetition_methods}
\end{table*}

\begin{table}[htbp]

\centering
\begin{adjustbox}{max width=\textwidth/2}
\begin{tabular}{lcccccccc}
\toprule
\textbf{Method} & \textbf{rep-2↓} & \textbf{rep-3↓} & \textbf{rep-w↓} & \textbf{rep-r↓} & \textbf{div↑} & \textbf{uniq-1↑} \\
\midrule
Ori.           & 93.29           & 92.65           & 0.91           & 0.95           & 0.00            & 7355           \\
CE          & 6.53           & 6.27            & 0.06            & 0.04            & 0.82           & 7190            \\
UN          & 3.88           & 3.44            & 0.06            & 0.03            & 0.89           & 6984           \\
CL         & 3.72           & 3.30           & 0.06            & 0.03           & 0.90            & 7155           \\
CT         & 0.59          & 0.10            & 0.04           & 0.02            & 0.99            & 7395            \\
CTSD       & 0.69           & 0.14            & 0.05            & 0.02           & 0.99           & 7104            \\
\bottomrule
\end{tabular}
\end{adjustbox}
\caption{Top 1\% repeatability metrics among 1 million items, titles from online e-commerce websites.}
\label{table:million}
\vspace{-0.5em}
\end{table}

The research shows that CT loss is the optimal loss function in suppressing oscillations hallucination, as it only suppresses negative tokens while enhancing positive tokens. Despite its effectiveness, its rough selection of negative tokens sometimes leads to suboptimal results. Therefore, based on the previously analyzed repeatability principle of ALTI+ and T-SNE, this paper proposes CTSD loss, an optimization of the original CT loss, significantly improving the accuracy, stability, and effectiveness of the output of the new training model. This method dynamically suppresses previously generated tokens by designing two attenuation factors. The first attenuation factor uses cosine similarity to measure the similarity in the context that the hallucination token's attention is very similar to the previous token. Additionally, an exponential-decay attenuation factor is designed to weaken the suppression of distant tokens, considering that the contribution of generated tokens is inversely related to the distance between tokens.

Finally, CTSD loss for time step \(t\) is defined as:
\vspace{-0.5em}
\begin{equation}
\small
\mathcal{L}_{CTSD}^t= \log \left(1+\!\sum_{y_t^{-} \in S_N^t}\alpha_d \alpha_s \exp \left(h_t^T W_{y_t^{-}}\! -h_t^T W_{y_t}\right)\right)
\end{equation}
where \(\alpha_{d} = e^\frac{{t_{-}-t} }{T} \), \(t_{-}\) represents the time when \(y_t^{-} \) is generated, \(T \) is the temperature coefficient that controls decay. \(\alpha_{s} = \frac{atten_{t_{-}}^{\text{T} } atten_{t} }{\left \|atten_{t_{-}}  \right \|\left \|atten_{t}  \right \| } \), \(atten_{t_{-}}\) represents attention distribution between \(y_t^{-} \) and encoder embedding. 

\begin{figure}[htb]
    \centering
    \includegraphics[width=1\linewidth]{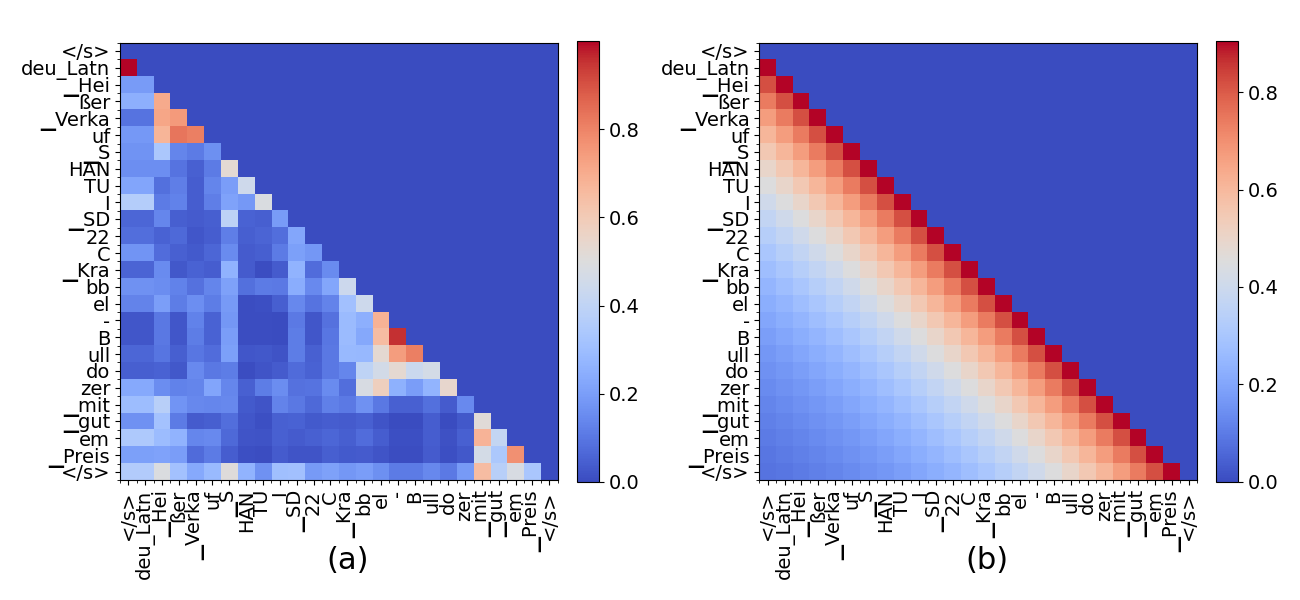}
    \caption{Attenuate factor of different generated tokens. (a) attention similarity and (b) exponential decay matrix.}
    \label{fig:attenuate_factor}
\vspace{-1.5em}
\end{figure}

To more intuitively demonstrate the role of the attenuation factor, we display the weight matrix of a normal translated sentence (constructed from the attenuation factor of each generated token) in Figure~\ref{fig:attenuate_factor}. Figure~\ref{fig:attenuate_factor}(a) is the attention similarity matrix. Figure~\ref{fig:attenuate_factor}(b) is the exponential decay matrix. It can be observed that under normal translation results, the overall attention similarity between tokens is low, but some unrelated tokens that are far apart also have high similarity. Hence, an additional exponential-decay attenuation factor must be added for secondary suppression.

\begin{table*}[htbp]
\centering
\begin{adjustbox}{max width=0.95\textwidth}
\renewcommand{\arraystretch}{0.92}
\begin{tabular}{lcccccccccc}
\toprule
\textbf{Model} & \textbf{Method} & \textbf{SacreBLEU↑} & \textbf{Rouge-L↑} & \textbf{COMET↑} & \textbf{rep-2↓} & \textbf{rep-3↓} & \textbf{rep-w↓} & \textbf{rep-r↓} & \textbf{div↑} & \textbf{uniq-1↑} \\
\midrule
\multirow{6}{*}{Qwen-7B}   & Ori.      & 4.14         & 0.133           & 0.599     & 5.66           & 5.22           & 0.04           & 0.03           & 0.85            & 14072            \\
 & ConvDPO     & 1.87         & 0.082           & 0.509     & 5.77           & 2.73           & 0.04           & 0.08           & 0.90            & 33832            \\
 & TransDPO          & 0.22             & 0.026               & 0.278              & 49.46           & 39.15            & 0.45            & 0.58            & 0.21           & 24292            \\
 & TransDPOLoRA        & 0.67             & 0.032               & 0.356               & 30.11           & 25.24           & 0.28            & 0.31           & 0.40            & 13702           \\
 & CovDPO+CE          & 21.53        & 0.419           & 0.711     & 0.75         & 0.12            & 0.05           & 0.02            & 0.99           &  10826            \\
 & CovDPO+CTSD      & 21.71             & 0.421             & 0.708              & 0.74           & 0.21            & 0.05            & 0.02            & 0.99     & 10778            \\
\bottomrule
\end{tabular}
\end{adjustbox}
\caption{Translation quality and repetition rate of Qwen-7B under DPO and training methods.}
\vspace{-0.95em}
\label{table:dpo}
\end{table*}

\section{EXPERIMENTS}

\subsection{Experiments Setup}

The experiments aim to evaluate whether CTSD can suppress hallucinations in specialized translation models and LLMs while maintaining stability. We integrate several baseline methods, including traditional CE loss, decoding-based methods like CS and PS, and training-based methods such as UL at Token-level (UL-T), CL, and CT.

For specialized models like NLLB-1.3B and mBART-large. Through extensive comparative analysis and experimentation, a batch size of 64 and a fixed learning rate of \(5 \times 10^{-5}\) can effectively balance stability and ultimate performance. For decoder-only LLMs, like LLaMA2-7B and Qwen-7B, the LoRA method with parameters r and alpha set to 8 and 16, respectively, is applied. The batch size for these models is set to 32, and a learning rate is established at \(2 \times 10^{-5}\), aimed at consistently improving the translation capabilities of LLMs. The hyperparameter ablation experiment for the CTSD method is discussed in the appendix.

Our datasets comprise the open-source general dataset WMT16 for training and the FLORES-200 devtest dataset for evaluation. Furthermore, a novel evaluation dataset comprising e-commerce texts susceptible to hallucination translations is compiled and released to benchmark our algorithm. This dataset is an English-German Parallel Corpus encompassing 3,500 authentic titles from alibaba.com. Each text segment has undergone meticulous translation and verification by human experts.

NMT Performance is evaluated by SacreBLEU, Rouge-L, and COMET, along with repeatability metrics such as 2-gram and 3-gram repetition rates (rep-2 and rep-3), token dispersion (div), rep-w, and rep-r. Rep-n measures sequence-level repetition, div estimates repetition at different n-gram levels, and rep-w calculates the proportion of current tokens occurring within the previous \(w\) tokens.

It is worth noting that lower repeatability metrics do not always indicate better NMT quality. 

\begin{table*}[tbp]
\centering
  \renewcommand{\arraystretch}{0.95}
  \begin{adjustbox}{max width=0.9\textwidth}
  \begin{tabular}{llcccccc}
    \toprule
    Comparison & Metric & AR & DE & TH & HI & HE & PT \\
     \midrule
    \multirow{2}{*}{CTSD vs. Baseline} & LQR-3 & +22.22\% & +6.91\% & +56.51\% & +85.12\% & +11.72\% & +40.22\% \\
                                       & LQR-4 & +54.10\% & +22.82\% & +156.96\% & +257.47\% & +44.89\% & +38.85\%  \\
    \midrule
    \multirow{2}{*}{CTSD vs. Google Translate} & LQR-3 & +9.66\% & +7.77\% & +22.81\% & +11.63\% & +8.28\% & +6.15\% \\
                                              & LQR-4 & +14.93\% & +0.10\% & +50.00\% & +3.33\% & +6.51\% & +38.45\%  \\
   \bottomrule
\end{tabular}
\end{adjustbox}
\caption{LQR-3, LQR-4 rates of CTSD compared to baseline and Google Translate under human evaluations.}
\label{tab:human_evaluation}
\vspace{-0.5em}
\end{table*}

\begin{table}[tbp]
\centering
  \begin{adjustbox}{max width=\textwidth/2}
  \begin{tabular}{lcccccc}
    \toprule   
    Metric & PV & UV & CTR & CVR & GMV & RPM \\
     \midrule	  
    AR & +0.74\% & +0.56\% & +0.38\% & +2.06\% & +2.96\% & +2.21\% \\
    DE & +0.67\% & +0.31\% & +0.44\% & +1.82\% & +0.26\% & +0.63\%  \\
   \bottomrule   
\end{tabular}
\end{adjustbox}
\caption{Online A/B testing results in www.alibaba.com}
\vspace{-0.9em}
\label{tab:onlineAB}
\end{table}

\subsection{Evaluation Results}

As shown in Table~\ref{table:fulltable}, CTSD consistently improves translation quality across all models, as evidenced by SacreBLEU, Rouge-L, and COMET metrics, while maintaining extremely low repetition rates.

For specialized translation models, the CT loss underperforms CE loss in the non-hallucination dataset, while CTSD significantly enhances performance on both e-commerce hallucination and general datasets. NLLB-1.3B and mBART-large showed notable improvements of +13.1\% and +4.0\% in SacreBLEU and +4.7\% and +5.47\% in COMET, respectively, substantially reducing repetition rates. For LLMs prompted for translation tasks, CTSD demonstrated significant improvements, particularly on the e-commerce hallucination dataset. LLaMA2-7B achieved +10.76\% in SacreBLEU and +5.50\% in Rouge-L compared to the CT model. Additionally, closed-source models like ChatGPT and GPT-4 scored lower in SacreBLEU but acceptable in COMET, with decent translation capabilities and strong hallucination suppression, while showing weaker professionalism for specialized tasks.

Experiments with the Qwen-1.8B and Qwen-14B models (Table~\ref{table:different scale model}) show that CTSD effectively maintains translation accuracy across different LLM sizes, emphasizing its robust enhancement of LLM translation capabilities regardless of hallucination tendencies.

To further verify the effectiveness of CTSD as a repetition suppression method, we conducted experiments comparing different methods, summarized in Table~\ref{table:repetition_methods}. Although the decoding method significantly improved the hallucination dataset (811.27\% increase in SacreBLEU and 103.93\% in COMET), its translation quality still lagged behind training methods. Among the training methods, CTSD stands out on both specialized translation models and LLMs, maintaining a meager repetition rate, indicating it is a general and efficient repetition suppression method.

In order to demonstrate the benefits of CTSD for normal title translations, we translated approximately 1 million e-commerce titles on alibaba.com using models trained with different methods. By filtering the top 1\% of repeated titles through the rep-w metric, the final repetition rates of different models are shown in Table~\ref{table:million}. It is evident that CT and CTSD methods outperformed other baselines, with rep-2 decreasing by 846.38\% and rep-w by 100\%, respectively. Compared to CT, the repetition rate of CTSD is slightly higher, which aligns with the nature of word stacking in e-commerce titles. This demonstrates that CTSD can suppress oscillation hallucinations and preserve the natural repetitive characteristics of e-commerce titles in the meantime.

In our final experiment, we evaluated the DPO method's effectiveness in mitigating model oscillation hallucinations, as shown in Table~\ref{table:dpo}. ConvDPO used general preference data, while TransDPO and TransDPOLoRA used private-domain translation preference data. Within the dataset of translation preferences, "Chosen answer" represented authentic e-commerce translations, and "rejected answer" represented base model translations.

The results show that whether using generic or private-domain data, DPO fails to address oscillation effectively for e-commerce translations. Moreover, DPO followed by LoRA fine-tuning for sub-tasks is less effective than direct LoRA fine-tuning of NMT tasks. In summary, while DPO is commonly used for suppressing hallucinations in LLMs, it is ineffective against oscillation hallucinations in e-commerce translation contexts. Our results underline that CTSD is a superior solution for this specific challenge.

\subsection{Online E-commerce Experiments}
 
We implemented the CTSD algorithm on the specialized translation model of the www.alibaba.com website, which uses an encoder-decoder structure with 48 layers and approximately 1.1B parameters. We selected six high-traffic language websites (AR - Arabic, DE - German, TH - Thai, HI - Hindi, HE - Hebrew, and PT - Portuguese) to translate item titles and descriptions. These sites serve millions of users, generating nearly 20 times daily page views (PVs). A/B tests were conducted with models fine-tuned on an e-commerce dataset, comparing CTSD and non-CTSD models. Each user saw translation text from only one model to ensure fairness.

First, to ascertain the impact on online translation accuracy, we performed a pre-procedure expert evaluation. 2,000 items were randomly selected and translated by both models. Experts rated the translations on a 5-point scale,  with LQR-3 (or 4) indicating that at least two experts rated the translation more than 3 points (or 4 points). Table~\ref{tab:human_evaluation} shows that the CTSD-trained model significantly improved translations across all six languages, particularly for languages with fewer training data (AR, TH, HI, HE). Additionally, the CTSD model outperformed Google Translate overall, with the Fleiss Kappa mean value exceeding 0.6, demonstrating high consistency among raters.

For online evaluations, we assessed business indicators such as page view (PV), retained user (UV), click-through rate (CTR), average conversion rate (CVR), gross merchandise volume (GMV), and revenue per mille (RPM). The online A/B experiments in AR and DE (Table~\ref{tab:onlineAB}) showed that the new translation model improved title translation quality, leading to greater product attention and significant enhancements in all indicators, especially GMV and RPM, which enhanced by 2.96\% and 2.21\% on Arabic sites, respectively.

All data and code implements will be released to the public after publication.

\section{Conclusion}

In conclusion, this study addresses the critical challenge of repetition generation in NMT. By analyzing and visualizing the underlying causes from the lens of information entropy, we propose one novel method, which can dynamically modulate token suppression to reduce the redundancy of some generated words. Extensive experiments on offline general and e-commerce datasets and rigorous online A/B tests have verified its performance in improving translation quality and handling oscillation hallucinations.

\section{Limitations}
While our CTSD method has shown significant improvements in reducing repetition and enhancing translation quality, there are some limitations to consider.  Firstly, the optimal settings for temperature coefficient and decay factor may vary across models and datasets. Automatic tuning for these hyperparameters needs further investigation. Secondly, the additional computations for attention similarities and decay factors during training have not been fully analyzed. Assessing the trade-off between performance gains and computational costs is necessary, especially for resource-limited environments.  Addressing these limitations in future work can enhance the robustness and applicability of the CTSD method, contributing to more reliable NMT systems.

\section{Ethics Statement}
In this work, we employed publicly released and private e-commerce domain datasets to train our machine translation models. Public datasets have been carefully reviewed for ethical concerns, and our manual inspections found no significant ethical issues, such as violent or offensive content. The e-commerce datasets are anonymized and collected with proper consent, following data protection regulations. We also intend to share our source code with clear instructions to encourage ethical use. Despite these precautions, machine translation can sometimes produce unexpected outputs. We will implement mechanisms to reduce such risks and advise users to follow ethical guidelines to prevent misuse.

\bibliography{custom}

\appendix

\section{Explanation of Metrics}

\begin{table*}[h]
\centering
\caption{The impact of hyperparameters W, N, and T in CTSD on translation quality and repeatability.}
\begin{adjustbox}{max width=\textwidth}
\begin{tabular}{lcccccccccccc}
\toprule
\textbf{Model} & \textbf{Weight} & \textbf{PredToken} & \textbf{T} & \textbf{SacreBLEU↑} & \textbf{Rouge-L↑} & \textbf{COMET↑} & \textbf{rep-2↓} & \textbf{rep-3↓} & \textbf{rep-w↓} & \textbf{rep-r↓} & \textbf{div↑}  \\
\midrule
\multirow{8}{*}{NLLB-1.3B}   & 0.1      & 10         & 5           & 7.04     & 0.187          & 0.558           & 61.41           & 62.02           & 0.15       & 0.15      & 0.16     \\
 & 0.5          & 5           &  5             & 7.58         & 0.193           & 0.585     & 50.56           & 51.2           & 0.18           & 0.18           & 0.11                      \\
 & 0.5          & 5         & 10           & 7.78       & 0.197          & 0.595            & 43.86           & 44.61            & 0.15            &  0.15       & 0.17     \\
 & 1.0          & 2             & 5             & \textbf{8.19}              & 0.202           & 0.616            & 24.37            & 24.73            & 0.10            & 0.09   &  0.42           \\
 & 1.0          & 5             & 5               & 7.99              & 0.201           & 0.606            & 34.98          & 35.23            & 0.14           & 0.13    &  0.27       \\
 & 1.0         & 10             & 5               & 8.15              & 0.202           & 0.614           & 26.08            & 26.32           & 0.11   & 0.10          & 0.39             \\
 & 2.0          & 10             & 5               & \textbf{8.19}               & \textbf{0.203}          & \textbf{0.622}            & \textbf{12.66}           & \textbf{12.21}            & \textbf{0.07}            & \textbf{0.06}   & \textbf{0.67}         \\

\midrule 
\multirow{8}{*}{Qwen-7B}   & 0.005      & 10         & 5           & 23.93     & 0.451           & 0.737           & \textbf{0.62}          & \textbf{0.09}           & \textbf{0.04}            & 0.02  & 0.99            \\
 & 0.01     & 5         & 5           & 23.70     & 0.457          & \textbf{0.740}          & 0.73           & 0.14           & 0.05            & 0.02   & 0.99            \\
 & 0.01          & 5        & 10           & \textbf{24.35}     & 0.457        & \textbf{0.740}            & 0.75         & 0.21           & 0.05           &  0.02   & 0.99            \\
 & 0.02          & 5             & 5             & 24.22              & \textbf{0.460}          & 0.739            & 0.72            & 0.13            & 0.05       & 0.02     & 0.99            \\
 & 0.02          & 10             & 5               & 23.97              & 0.459           & 0.738            & 0.67            & 0.12            & 0.05           & 0.02     & 0.99            \\
 & 0.02         & 20             & 5               & 22.64               & 0.437           & 0.728           & 0.78            & 0.22           & 0.05        & 0.02    & 0.99           \\
 & 0.1          & 10             & 5               & 22.37             & 0.432               & 0.717               & 3.84          & 3.48           & 0.05           & 0.02           & 0.89                 \\
\bottomrule
\end{tabular}
\end{adjustbox}
\label{table:hyperparameters}
\end{table*}

In this paper, rep-w is calculated by the proportion of current tokens occurring within the previous \(w\) tokens, expressed as:
\begin{equation}
\begin{aligned}
\small
\text{rep-w} = \frac{1}{|\mathcal{D}|} \sum_{s \in \mathcal{D}} \frac{1}{|s|} \sum_{t=1}^{|s|} \mathbf{1}\left[s_t \in s_{t-w-1: t-1}\right]
\end{aligned}
\end{equation}
where \(\mathcal{D}\) represents the result set, s represents generated sentences in \(\mathcal{D}\). Moreover, rep-r stands for the ratio of the repetition snippet in a sentence measured by length, defined as:

\begin{equation}
\small
\begin{aligned}
\text{rep-r} = \frac{1}{|s|} \Biggl|\Biggl\{i \mid & \left(s_i=s_j \wedge s_{i+1}=s_{j+1}, \exists j \neq i\right) \\
& \vee \left(s_i=s_k \wedge s_{i-1}=s_{k-1}, \exists k \neq i\right)\Biggr\}\Biggr|
\end{aligned}
\end{equation}
\section{Additional Experimental Results}

\begin{figure}
    \centering
    \includegraphics[width=1\linewidth]{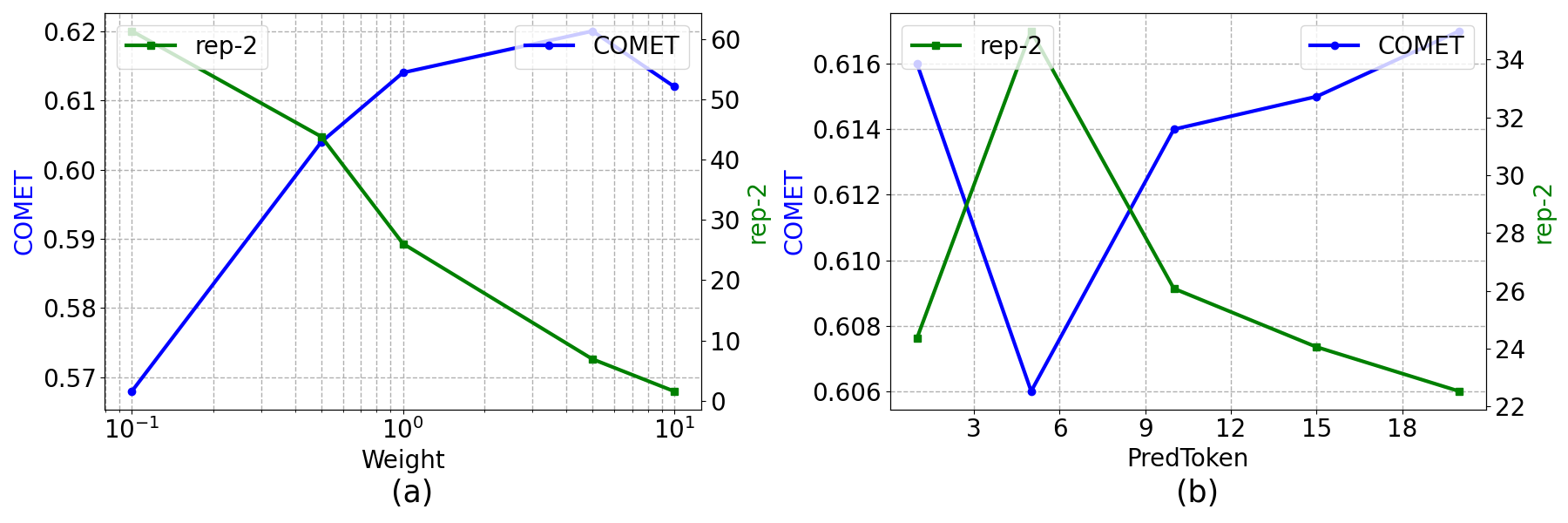}
    \caption{The impact of hyperparameters W and N on the translation quality and reproducibility of the NLLB-1.3B model. }
    \label{fig:nllb}
\end{figure}

\begin{figure}
    \centering
    \includegraphics[width=1\linewidth]{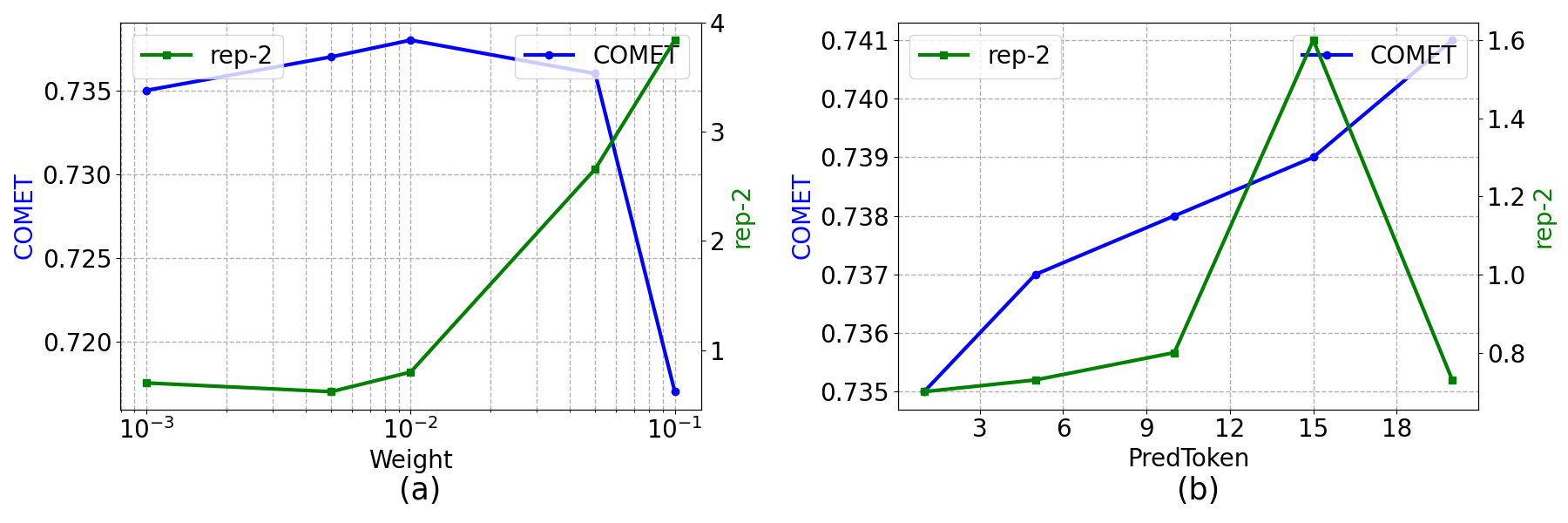}
    \caption{The impact of hyperparameters W and N on the translation quality and reproducibility of the Qwen-7B model. }
    \label{fig:qwen}
\end{figure}

We discuss the impact of the three hyperparameters W, N, and T within the CTSD algorithm on the translation quality and repetition performance across various models, where W represents the ratio of CE loss with CTSD loss, N denotes the number of previous tokens to focus on, and T represents the temperature coefficient of \(\alpha_{d}\).

The analysis presented in Table ~\ref{table:hyperparameters} and Figure ~\ref{fig:nllb} demonstrates that for specialized translation models like NLLB-1.3B, a moderate increase in T can lead to an improvement in translation quality with repetition rates exhibiting the opposite trend. On the other hand, when W increases, the translation quality first increases and then decreases while the repetition rate continues to decrease. However, as N changes, there is an initial decrease in translation quality followed by an increase that improves and then deteriorates, and the repetition rate consistently declines. These results suggest that incorporating a suitable CTSD loss into specialized model training and setting a larger window for it can effectively reduce repetition rates and enhance model performance. However, it is important to note that an excessively high weight on the CTSD loss can disrupt the original training direction of the model. Moreover, larger windows require increased computational demands, which poses a trade-off between accuracy and training duration. However, the results for Qwen-7B exhibit a distinct pattern. As shown in Figure ~\ref{fig:qwen}, when W increases, the repetition rate initially decreases and then rises. This implies that increasing CTSD loss continuously during large model training not only diminishes the translation quality but also induces new oscillatory hallucinations.
\label{sec:appendix}

\end{document}